\documentclass[conference]{IEEEtran}
\IEEEoverridecommandlockouts

\usepackage{cite}
\usepackage{amsmath,amssymb,amsfonts}
\usepackage{algorithmic}
\usepackage{graphicx}
\usepackage{textcomp}
\usepackage{xcolor}
\def\BibTeX{{\rm B\kern-.05em{\sc i\kern-.025em b}\kern-.08em
    T\kern-.1667em\lower.7ex\hbox{E}\kern-.125emX}}

\usepackage{url}
\usepackage[hidelinks]{hyperref}
\usepackage{pifont}
\usepackage{siunitx}
\usepackage{booktabs}
\usepackage{graphicx}
    
\begin{document}

\title{Entropy-Driven Curriculum for Multi-Task Training in Human Mobility Prediction
\thanks{This work was supported by the German Federal Ministry of Research, Technology and Space (BMFTR) under grant number 16DHBKI020.}
}

\author{
\IEEEauthorblockN{Tianye Fang}
\IEEEauthorblockA{
\textit{Technical University of Munich}\\
Munich, Germany \\
tianye.fang@tum.de}
\and
\IEEEauthorblockN{Xuanshu Luo}
\IEEEauthorblockA{
\textit{Technical University of Munich}\\
Munich, Germany \\
xuanshu.luo@tum.de}
\and
\IEEEauthorblockN{Martin Werner}
\IEEEauthorblockA{
\textit{Technical University of Munich}\\
Munich, Germany \\
martin.werner@tum.de}
}

\maketitle

\begin{abstract}
The increasing availability of big mobility data from ubiquitous portable devices enables human mobility prediction through deep learning approaches. However, the diverse complexity of human mobility data impedes model training, leading to inefficient gradient updates and potential underfitting. Meanwhile, exclusively predicting next locations neglects implicit determinants, including distances and directions, thereby yielding suboptimal prediction results. This paper presents a unified training framework that integrates entropy-driven curriculum and multi-task learning to address these challenges. The proposed entropy-driven curriculum learning strategy quantifies trajectory predictability based on Lempel-Ziv compression and organizes training from simple to complex for faster convergence and enhanced performance. The multi-task training simultaneously optimizes the primary location prediction alongside auxiliary estimation of movement distance and direction for learning realistic mobility patterns, and improve prediction accuracy through complementary supervision signals. Extensive experiments conducted in accordance with the \textit{HuMob Challenge} demonstrate that our approach achieves state-of-the-art performance on GEO-BLEU (0.354) and DTW (26.15) metrics with up to 2.92-fold convergence speed compared to training without curriculum learning.
\end{abstract}


\section{Introduction}

The inherent regularity of human mobility data, which exhibits predictability of individual mobility patterns across diverse populations and travel distances \cite{song2010limits}, provides the foundation for numerous location-based applications, including urban planning and management, transportation optimization, epidemic modeling, and recommendation systems \cite{7840811, 10.1177/23998083221075634, 10.5555/3060832.3060987, Pappalardo2023, 7373330, 9861127}. With the proliferation of pervasive user devices with passive location acquisition capabilities, unprecedented volumes of human mobility data have been collected, enabling data-driven approaches, particularly sequential deep learning models, to effectively extract human mobility patterns \cite{Liu_Wu_Wang_2016, Jiang_Song_Fan_2018, 10.1145/3178876.3186058, ijcai2018p324}. In comparison to handcrafted pattern matching \cite{10.1145/2501654.2501656, 10.1145/2483669.2483682, 10.1145/3615894.3628501} and Markov models \cite{QIAO201899, 8812927, 9430503}, deep learning methods generally achieve superior long-term prediction performance.

\begin{figure}[t]
    \centering
    \includegraphics[width=0.905\columnwidth]{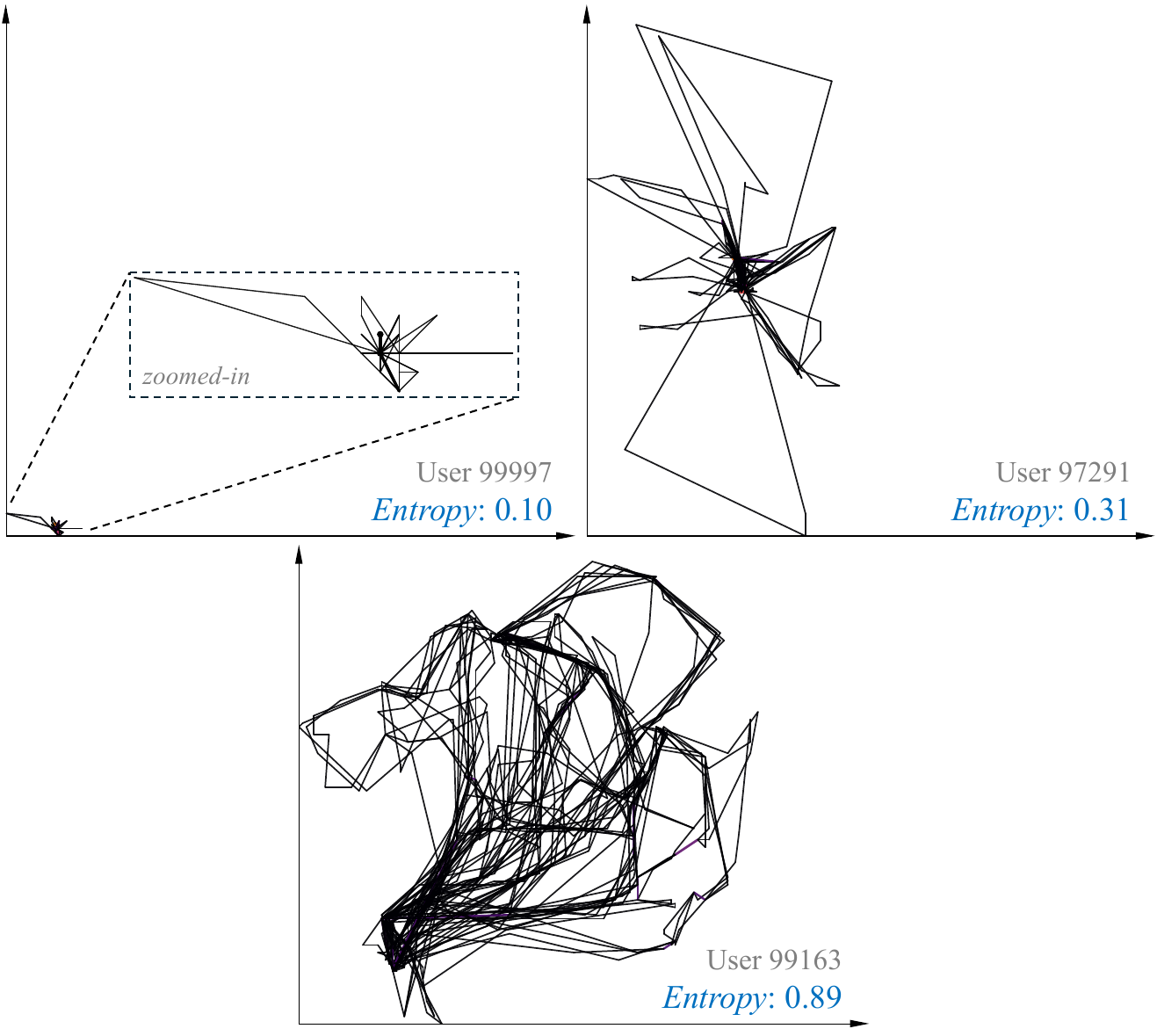}
    \caption{Three trajectories with heterogeneous complexity in the \texttt{YJMob100K} dataset \cite{Yabe2024}, presented at the \textbf{same} spatial scale. The entropy reflects trajectory predictability, which will be explained in Section \ref{sec:entropy1} and \ref{sec:entropy2}.}
    \label{fig:entropy_comp}
\end{figure}

Training these deep learning models conventionally involves random dataset shuffling and uniform mini-batch sampling, thereby implicitly treating all mobility data as equally difficult to learn, which conflicts with the intrinsic heterogeneous complexity of human mobility data, as illustrated in Fig. \ref{fig:entropy_comp}. Specifically, simple, repetitive routines (e.g., daily commutes) are fundamentally easier to model than irregular, sparse trajectories and complex trips with multiple stops. Consequently, such training procedures force models to learn from intricate examples early when the foundational understanding of mobility patterns remains undeveloped, leading to inefficient learning, training instability, premature convergence, and degraded prediction performance. Therefore, presenting human mobility data in order of increasing difficulty constitutes a rational training strategy, aligning with the concept of curriculum learning \cite{9392296, NIPS2010_e57c6b95, 10.1145/1553374.1553380}. However, to the best of our knowledge, curriculum learning has not been extensively applied to human mobility prediction. Moreover, there is a lack of practical and validated methods to quantitatively evaluate trajectory complexity for data reordering in curriculum organization.

Furthermore, most human mobility prediction approaches predominantly focus on next-location prediction as a single training objective while neglecting other inherent mobility characteristics that could provide valuable supervision signals. Concretely, human mobility inherently involves multifaceted decision-making processes where one may simultaneously consider destinations, means of transportation, and directional preferences based on urban structure or personal routines. By optimizing solely for location accuracy, models fail to capture these complementary aspects of mobility behaviors. Therefore, integrating multiple training objectives through multi-task learning (MTL) presents a promising approach to leverage these auxiliary mobility characteristics and enhance the primary next-location prediction \cite{Caruana1997}, enabling models to learn more comprehensive and realistic representations of human mobility. Although several studies have utilized MTL and obtained superior prediction performance \cite{10.1145/3178876.3186058, 9130086, LEI2025102597}, their auxiliary task semantics are context-specific, depending heavily on underlying datasets. For example, the next activity \cite{10.1145/3178876.3186058, 9130086} and transportation modes \cite{LEI2025102597} are not universally available in human mobility datasets, leading to limited generalization capabilities to other datasets and tasks.

To address these issues, this paper presents a unified training framework that integrates entropy-driven curriculum and MTL for human mobility prediction. Concretely,
\begin{itemize}
    \item The proposed entropy-driven curriculum learning strategy quantifies trajectory predictability by estimating mobility entropy based on the Lempel-Ziv (LZ) compression principle \cite{1055672, 1055714}. Training is then organized using data with increasing mobility entropy and prediction horizon, incorporating trajectory augmentation through mirroring and rotation, followed by fine-tuning exclusively on real trajectories for adaptation to original datasets.
    \item The primary location prediction task is jointly optimized alongside two universally applicable auxiliary estimations on moving distances and directions, which are inherently present in any mobility dataset and provide complementary supervision signals for more comprehensive mobility representations without additional data annotations.
\end{itemize}
Regarding the underlying model in the proposed framework, we present MoBERT, an encoder-only Transformer designed for human \underline{mo}bility prediction based on \underline{BERT} \cite{NIPS2017_3f5ee243, devlin-etal-2019-bert}, which utilizes multi-feature embeddings, including temporal, spatial, and potential semantic information, coupled with feature interaction mechanisms to capture complex dependencies. We evaluate our method on \texttt{YJMob100K} \cite{Yabe2024}, an open-source urban-scale human mobility dataset utilized in the \textit{HuMob Challenge}\cite{10.1145/3615894, 10.1145/3681771}, with standardized data splits and evaluation metrics: GEO-BLEU \cite{10.1145/3557915.3560951} and dynamic time warping (DTW) \cite{1104847}. We adhere precisely to the same settings as the \textit{HuMob Challenge} and demonstrate that our approach achieves state-of-the-art GEO-BLEU score of 0.354 and DTW distance of 26.15, with up to 2.92-fold convergence speed compared to training without the entropy-driven curriculum. For previously unseen cities, our method also exhibits superior zero-shot generalization capability. These results validate the effectiveness of the proposed framework for human mobility prediction. The code of this work is available here\footnote{\url{https://github.com/tum-bgd/2025-IEEEBigData-HuMob}}.

\section{Background and Related Work}

\subsection{Human mobility prediction}

A human mobility trajectory is a temporally ordered sequence of spatiotemporal points that captures an individual's movement over time. Specifically, let a $j$-step trajectory be represented as $\mathbf{T}_{1:j} = \{p_1, \ldots, p_j\}$, where each point $p_i$ contains at minimum a spatial coordinate $l_i = (x_i, y_i)$ and a corresponding timestamp $t_i$. Human mobility prediction aims to forecast individuals' future locations and movement patterns based on their historical trajectory data. Given an observed trajectory prefix of length $n$, denoted as $\mathbf{T}_{1:n}$, the objective is to predict the subsequent trajectory segment $\mathbf{T}_{n+1:n+k}$, where $k$ is the prediction horizon, by maximizing the conditional probability $P(\mathbf{T}_{n+1:n+k} \ | \ \mathbf{T}_{1:n})$. The prediction accuracy depends on capturing the underlying spatiotemporal dependencies and behavioral patterns encoded in the observed trajectory. Spatial coordinates may be continuous geographical coordinates such as longitude and latitude, or discrete grid indices in rasterized spaces, making the prediction a regression or classification task, respectively. Meanwhile, each point $p_i$ may also comprise semantic information $s_i$, e.g., the functional category of $l_i$, providing additional contribution to prediction.

Early approaches to human mobility prediction relied on either handcrafted rules \cite{10.1145/2501654.2501656, 10.1145/2483669.2483682, 10.1145/3615894.3628501} or Markov models \cite{QIAO201899, 8812927, 9430503}, where the latter treats location transitions as probabilistic state changes. However, these methods are generally insufficient to characterize the complexity of human mobility and capture the nonlinear dependencies and long-range patterns inherent in human movement \cite{9204396}. On one hand, human movement exhibits complex periodic patterns, and actual trajectory transitions do not conform to the assumptions of simple Markov chains. On the other hand, individuals often display different travel preferences, resulting in group-based rules performing poorly at the individual level. Moreover, these methods typically exploit only a limited history of past states, making it difficult to capture long-term dependencies and demonstrating limited scalability to increasingly large datasets.

The emergence of deep learning and the availability of big mobility data have revolutionized human mobility prediction. Recurrent neural networks and their variants have become the dominant approaches for modeling sequential trajectory data \cite{Liu_Wu_Wang_2016, Jiang_Song_Fan_2018, 10.1145/3178876.3186058, ijcai2018p324}. These models naturally handle variable-length sequences and capture temporal dependencies through their recurrent structure. However, they suffer from gradient vanishing problems in long sequences and process data sequentially, thereby limiting parallelization efficiency. There are also methods utilizing convolutional neural networks to extract local spatial patterns \cite{10.1145/3219819.3219931, 9378429}, but they struggle with global spatial and temporal dependencies and lack generalization capability across different spatial areas. While graph neural networks excel at learning local spatial relationships, these models often lack explicit temporal modeling, and their global spatial understanding remains limited \cite{Liu_Rong_Guo_2023, WANG2024129872}. Transformers \cite{NIPS2017_3f5ee243} represent a more advanced approach by handling long-range spatiotemporal dependencies simultaneously. Although Transformers typically require considerably large datasets for proper training \cite{osovitskiy2021an}, this is not problematic given the progressively expanding volume of human mobility data \cite{Yabe2024}. The analysis above motivates us to design a Transformer-like model in our proposed framework.

\subsection{Curriculum learning}

Curriculum learning is a training strategy mimicking human education, where learning progresses from simple to complex examples rather than through random sampling \cite{NIPS2010_e57c6b95}. The core methodology of curriculum learning is organizing training data by difficulty and gradually introducing more challenging examples as model competence develops. This approach contrasts with conventional training that randomly samples mini-batches, treating all examples as equally difficult throughout the training process \cite{9392296}. Curriculum learning typically outperforms random sampling for several reasons. First, starting with simple examples helps establish robust feature representations and prevents the model from struggling with complex data early in training. Second, the gradual increase in difficulty creates a smoother optimization landscape, reducing the likelihood of becoming trapped in poor local minima. Third, this approach naturally implements a form of regularization, as the model must maintain performance on simple examples while adapting to complex ones, thereby preventing overfitting to edge cases \cite{10.1145/1553374.1553380}.

Unfortunately, to the best of our knowledge, curriculum learning has rarely been applied to human mobility prediction despite the prevalent heterogeneous complexity of human mobility data, as illustrated in Figure \ref{fig:entropy_comp}. One marginally relevant work is \textit{ST-Curriculum Dropout} for spatial-temporal graph modeling to forecast traffic speed and taxi demand \cite{Wang_Chen_Pan_2023}, which measures vertex difficulty according to its latent representation. This approach relies heavily on rational model initialization and is susceptible to variations during training, potentially inducing misleading curricula that deteriorate training performance. Therefore, this paper focuses on data itself, formulating mobility entropy as a feasible measure of trajectory predictability independent of underlying models, thereby providing a reliable foundation for curriculum organization.

\subsection{Multi-task learning}

Multi-task learning (MTL) is a paradigm where a model simultaneously learns multiple related tasks, sharing representations to improve generalization across all objectives \cite{Caruana1997}. Unlike single-task learning (STL), which optimizes for one specific target, MTL leverages task relationships through parameter sharing, typically implemented via shared encoder layers with task-specific heads. This architecture enables the model to learn more robust features by considering multiple aspects of the data simultaneously. MTL generally outperforms STL for following reasons. First, auxiliary tasks act as regularizers, preventing overfitting to the main task by introducing additional training signals. Second, shared representations must capture features relevant to all tasks, leading to more generalizable learned patterns. Third, MTL effectively increases the amount of training data per shared parameter, as gradients from multiple tasks contribute to updating common layers. MTL is also beneficial when labeled data for the main task is limited, as related tasks can provide extra supervision.

In the context of human mobility prediction, the sparse and noisy nature of real-world mobility data exacerbate the challenge of learning robust representations from a single supervisory signal. Consequently, numerous methods have integrated MTL to optimize the primary mobility prediction task alongside activity type classification \cite{10.1145/3178876.3186058, 9130086} or transportation mode recognition \cite{LEI2025102597}. However, these auxiliary tasks require additional annotations that may not be universally available. In contrast, our approach introduces distance and direction prediction as auxiliary tasks, as elaborated in Section \ref{sec:model}, which can be derived from any trajectory dataset without additional labeling, thereby ensuring broad applicability while maintaining the benefits of MTL for enhanced accuracy.

\section{Methodology}\label{sec:methodology}

This section aims to elucidate the proposed unified training framework. We first employ Fano's inequality \cite{335943} to demonstrate that low-entropy trajectories are fundamentally more learnable, thereby establishing the theoretical foundation for entropy-driven curriculum organization. Subsequently, the normalized Lempel–Ziv mobility entropy estimator is proposed to quantify the predictability of mobility data. The entropy-driven curriculum learning procedure is then introduced along with trajectory augmentation strategies, followed by the design details of MoBERT, a BERT-based encoder-only Transformer for human mobility prediction that supports MTL.

\subsection{The relationship between entropy and predictability}\label{sec:entropy1}

The relationship between entropy and predictability in human mobility is grounded in information theory. Entropy measures the uncertainty or randomness in a sequence of observations, where higher entropy indicates greater unpredictability. For human mobility, trajectories with low entropy correspond to highly regular movement patterns, such as repetitive commuting routes between home and workplace. Conversely, high-entropy trajectories reflect more exploratory or irregular behaviors, such as tourist activities spanning diverse urban areas. The connection between mobility entropy and predictability limits can be established by Fano's inequality \cite{335943}. For a discrete random variable $X$ with alphabet size $Q$ and entropy $H(X)$, Fano's inequality provides a lower bound on the probability of prediction error $P_e$:
$$H(X) \leq H(P_e) + P_e \log_2(Q-1),$$
where $H(P_e) = -P_e \log_2 P_e - (1-P_e) \log_2(1-P_e)$ is the binary entropy of the error probability. Alternatively, let $\Phi = 1 - P_e$ be the maximum achievable prediction accuracy, we can rearrange the above inequality, yielding:
$$H(X) \leq -\Phi \log_2 \Phi - (1-\Phi) \log_2(1-\Phi) + (1-\Phi) \log_2(Q-1).$$
This relationship shows an upper bound on entropy given a certain level of predictability. The equality holds only when the prediction is optimal. In the context of human mobility prediction, $X$ represents the next location choice, $Q$ corresponds to the total number of possible locations in a rasterized area, and $H(X)$ measures the uncertainty in location selection. When entropy approaches its maximum value $\log_2 Q$, the prediction accuracy cannot exceed random guessing ($\Phi \approx 1/Q$). This occurs when movement patterns are completely random, with each location equally possible for the next step regardless of history. Conversely, when entropy approaches zero, the bound allows for near-perfect prediction ($\Phi \approx 1$). This scenario corresponds to highly deterministic movement patterns where future locations are totally determined by past behavior. Therefore, trajectories with low entropy are not only more regular but also fundamentally more learnable, rendering them suitable for early training phases and vice versa.

\subsection{The normalized Lempel–Ziv mobility entropy}\label{sec:entropy2}

The above theoretical foundation distinguishes the proposed entropy-driven method from heuristic difficulty measures, providing a principled approach grounded in information theory. In practice, we propose a normalized Lempel–Ziv mobility entropy estimator derived from Lempel–Ziv (LZ) compression \cite{1055672, 1055714}. Our approach approximates mobility entropy by analyzing the rate at which new subsequences appear in the trajectory, specifically encompassing the following steps:
\begin{enumerate}
    \item \textit{Trajectory symbolization}: Two-dimensional (2D) coordinates $l=(x, y)$ are flattened into 1D through the transformation $x \cdot (y_{max} + 1) + y$, where $y_{max}$ represents the maximum column index (counted from 0) in the spatial grid. This transformation preserves spatial relationships while facilitating sequential analysis.
    
    \item \textit{LZ parsing}: Symbolized trajectories are processed from left to right, maintaining a dictionary $\mathcal{D}$ of observed subsequences. At each position $m$, the shortest subsequence of length $q$ that has not previously appeared in $\mathcal{D}$ is recorded as a phrase, added to $\mathcal{D}$, and parsing continues from position $m+q$. The process continues until the end of the sequence.

    \item \textit{LZ entropy estimation}: The entropy is computed as:
    $$H_{\textit{LZ}} = \frac{\ln N}{\bar{Q} \ln 2},$$
    where $N = \sum_{u=1}^{\lvert\mathcal{D}\rvert} Q_u$, $\bar{Q} = N/\lvert\mathcal{D}\rvert$ is the average phrase length, and $\lvert\mathcal{D}\rvert$ denotes the number of phrases in $\mathcal{D}$. Longer $\bar{Q}$ indicates higher compressibility and lower entropy, reflecting more predictable movement patterns, whereas shorter $\bar{Q}$ suggests the converse \cite{1055714}.

    \item \textit{Normalization}: $H_{\textit{LZ}}$ is normalized for comparability across trajectories of different lengths by
    $$H_{\textit{norm-LZ}} = \frac{H_{\textit{LZ}}}{\log_2 N} \in [0, 1].$$
\end{enumerate}
$H_{\textit{LZ}}$ is normalized by dividing by $\log_2 N$, which is the maximum possible entropy for a fully random sequence, bounding the entropy measure between zero and one, where values approaching one indicate near-random mobility patterns, and values approaching zero suggest highly predictable behaviors.


\begin{figure}[t]
    \centering
    \includegraphics[width=0.75\columnwidth]{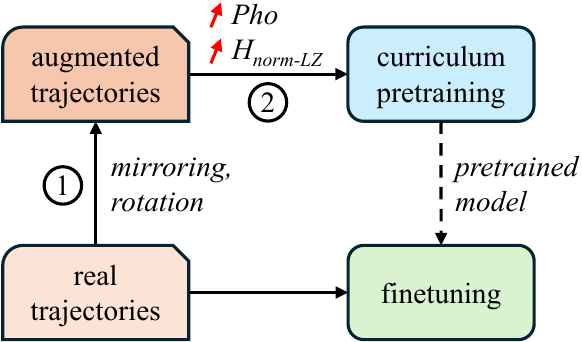}
    \caption{The entropy-driven curriculum learning pipeline. \ding{172} Real trajectories are augmented by mirroring and rotation. \ding{173} Augmented trajectories are ordered with increasing $H_{\textit{norm-LZ}}$ and subjected to progressively longer prediction horizon (\textit{Pho}) to form a curriculum for pretraining. The pretrained model is subsequently finetuned using the raw dataset.}
    \label{fig:curriculum}
\end{figure}

\begin{figure}[t]
    \centering
    \includegraphics[width=1.0\columnwidth]{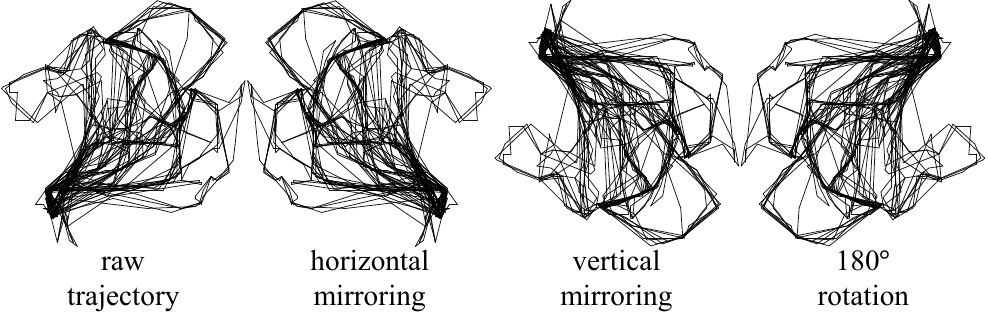}
    \caption{Trajectory augmentation by mirroring and rotation.}
    \label{fig:augmentation}
\end{figure}

\subsection{Entropy-driven Curriculum Learning}\label{sec:curriculum}

Building upon the normalized Lempel-Ziv mobility entropy $H_{\textit{norm-LZ}}$,  the entropy-driven curriculum learning framework organizes training progression according to intrinsic trajectory complexity, with the complete pipeline illustrated in Fig. \ref{fig:curriculum}.

Real trajectories are first augmented through horizontal and vertical mirroring, as well as \ang{180} rotation, quadrupling data volume while preserving underlying movement logic. Associated semantic information should also be transformed correspondingly to maintain consistency. Following augmentation, we perform entropy estimation of all trajectories by $H_{\textit{norm-LZ}}$, with trajectories sorted by increasing entropy, creating a natural progression from highly predictable to complex behaviors. To further enrich training diversity, sorted trajectories are assigned increasing \underline{p}rediction \underline{ho}rizon (\textit{Pho}). For example, given a task requiring forecasting of the next five locations of a trajectory ($\textit{Pho}=5$), variant samples can be created with $\textit{Pho}=1,2,\ldots,5$. The curriculum is then constructed by organizing augmented trajectories with increasing $H_{\textit{norm-LZ}}$ and $\textit{Pho}$, establishing progressive training difficulties.

This curriculum enables incremental learning, beginning with easily learnable patterns and gradually introducing more challenging scenarios. The entropy-based organization aligns training difficulty with information-theoretic predictability constraints, preventing early-phase overwhelm from complex patterns. Following curriculum pretraining, the model undergoes finetuning exclusively on real trajectories for optimal adaptation to authentic mobility characteristics. This two-stage pipeline balances large-scale pretraining on diverse augmented data with real-world adaptation, providing a principled approach that leverages entropy-based complexity measures for structured learning progression and improved generalization compared to conventional random sampling.

\begin{figure}[t]
    \centering
    \includegraphics[width=0.85\columnwidth]{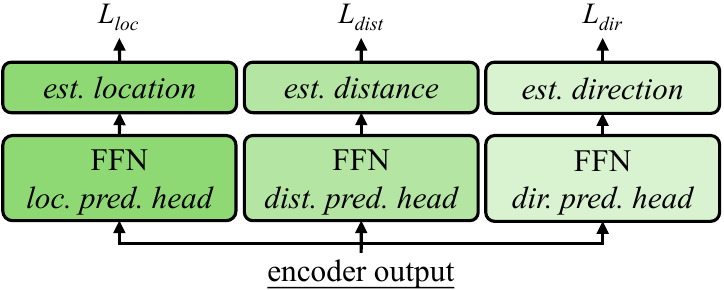}
    \caption{The multi-task prediction head. Losses are computed by cross-entropy.}
    \label{fig:multitask}
\end{figure}

\subsection{Model design}\label{sec:model}

To support MTL, this paper presents MoBERT, a BERT-like encoder-only model for human mobility prediction \cite{devlin-etal-2019-bert}. We adopt a BERT-style architecture for the following advantages. Unlike autoregressive models that generate predictions sequentially, BERT's bidirectional encoding and parallel prediction capability prevent error accumulation in long-term forecasting. The self-attention mechanism naturally captures both periodic patterns (e.g., daily commutes) and irregular dependencies without the computational constraints of recurrent architectures. Moreover, the masked language modeling paradigm of BERT aligns well with the sparsity of mobility data, where trajectories are often missing or irregularly sampled \cite{10.1145/3178876.3186058}.

The encoder part of MoBERT follows the architecture of the original BERT, where the input is adapted for human mobility data. Concretely, the encoder processes 3D input with dimensions $[B, M, E]$, where $B$ denotes the batch size, $M$ represents the length of historical trajectories, and $E$ indicates the embedding dimension. The specific composition of $E$ depends on the semantics provided in the dataset and minimally includes locations and corresponding timestamps.

For auxiliary tasks, we introduce distance and direction prediction, which are naturally available in any mobility dataset, avoiding dependence on specific datasets and ensuring the universality of MoBERT. These two straightforward tasks provide intrinsic spatial constraints: distance bounds the search space while direction provides orientation priors that complement the location prediction. In practice, both tasks can be formulated as classification or regression, depending on the dataset complexity and specific requirements. The prediction head of MoBERT is designed accordingly, as illustrated in Fig. \ref{fig:multitask}. They share the same latent representations through the encoder while employing task-specific feed-forward networks (FFNs) \cite{NIPS2017_3f5ee243}. We have the total loss
$$L = L_{loc} + \lambda_1 L_{dist} + \lambda_2 L_{dir},$$
where $L_{loc}$, $L_{dist}$, and $L_{dir}$ represents the losses for location, distance, and direction estimation, respectively, and $\lambda_1$, $\lambda_2$ denote the loss weights for auxiliary tasks. The values of $\lambda_1$ and $\lambda_2$ are determined via grid search within the range $[0, 1]$.

\section{Dataset and Adaptations}

To validate the effectiveness of the proposed method, this paper utilizes the \texttt{YJMob100K} dataset \cite{Yabe2024}, a large-scale, real-world human mobility dataset jointly developed by multiple research institutions, serving as one of the most comprehensive open-source trajectory datasets available for academic research. This dataset was employed in the \textit{HuMob Challenge} in 2023/24 \cite{10.1145/3615894, 10.1145/3681771}, with standardized train/test splits and evaluation metrics including GEO-BLEU \cite{10.1145/3557915.3560951} and DTW \cite{1104847}, facilitating transparent and fair comparison with existing studies that participated in the challenge. This section first briefly introduces the dataset, followed by the adaptations made to this dataset for the subsequent evaluations.

\subsection{The \texttt{YJMob100K} dataset}\label{sec:yjmob}

The \texttt{YJMob100K} dataset captures over 110 million trajectory records through smartphone-based high-frequency GPS signals collected from approximately 100,000 active users within a major urban area of Japan during a continuous 75-day observation period, where the last 15 days are used for prediction. All raw geographic information was anonymized and discretized into a uniform $200 \times 200$ grid, where each cell measures 500 meters per side. Temporal data are similarly discretized into 48 uniform time slots of 30 minutes each, encompassing complete daily cycles. Each trajectory record consists of five key fields: an anonymous user identifier \textsl{uId}, day index \textsl{dId} ranging from 0 to 74, time slot index \textsl{tId} from 0 to 47, and spatial coordinates $(x, y)$ indicating grid cell position. The dataset exhibits significant sparsity, reflecting realistic mobile phone usage patterns where users are not continuously tracked. The dataset includes comprehensive point-of-interest (\textsl{POI}) information for enhanced semantic understanding. Each grid cell contains counts for 85 anonymized \textsl{POI} categories covering dining, retail, education, healthcare, etc., enabling models to incorporate spatial semantics and urban context into mobility prediction tasks.

\subsection{Dataset-specific adaptations}\label{sec:adaptation}

The \texttt{YJMob100K} dataset provides rich semantic information from which the following features can be derived. The day index \textsl{dId} can be further formulated into the day of the week (\textsl{dayOfWeek}) to reflect weekly patterns. The time slot index \textsl{tId} can also be derived to obtain time intervals (\textsl{timedelta}) between consecutive trajectory points, which capture temporal continuity, and day/night features (\textsl{dayNight}) to distinguish between diurnal and nocturnal activities. For \textsl{POI}, only the top-\textit{k} most frequent \textsl{POI} categories, namely \textsl{topKPOI}, are considered, as most \textsl{POI} categories are rarely represented in the dataset. This approach emphasizes significant POIs while minimizing redundancy and noise, thereby enhancing the semantic representation of the environment. Consequently, the complete semantics $s$ of each sample in this dataset is represented as $s=\{\textsl{dayOfWeek}, \textsl{timedelta}, \textsl{dayNight}, \textsl{topKPOI}\}$, while the timestamp $t$ is defined as $t=\{\textsl{dId}, \textsl{tId}\}$. Combined with coordinates $(x, y)$, there are 8 features in total as input.

\subsubsection{Feature embedding interaction}\label{sec:interaction}

\begin{figure}[t]
    \centering
    \includegraphics[width=0.49\columnwidth]{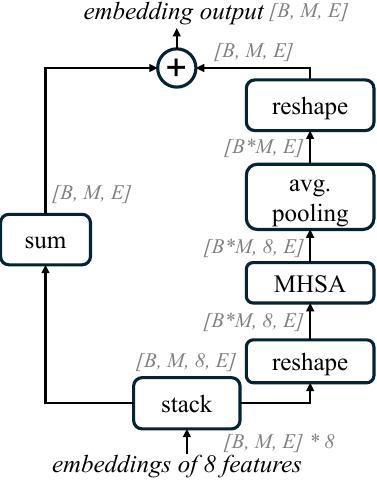}
    \caption{Feature interaction pipeline. The output shapes are denoted in gray.}
    \label{fig:interaction}
\end{figure}

To effectively utilize all 8 features, we design a feature interaction module based on multi-head self-attention (MHSA), effectively fusing features and providing input to the MoBERT encoder. All feature embeddings are first stacked and undergo basic fusion via summation over the feature dimension, as shown in the left branch in Fig. \ref{fig:interaction}. Subsequently, MHSA operates along the feature dimension to capture dynamic inter-dependencies, computing attention weights that adaptively emphasize different feature combinations based on context, for instance, prioritizing spatiotemporal interactions during commuting hours or \textsl{POI}-temporal relationships during leisure periods. The final representation combines both the basic fusion term (left) and the attention-based interaction term (right) via element-wise summation, preserving individual feature semantics while incorporating learned relational patterns that enhance the model's ability to understand complex mobility behaviors. The results in Section \ref{sec:results} validate the effectiveness of this module.

\subsubsection{Curriculum stages}\label{sec:curriculum-stages}

\begin{figure}[t]
    \centering
    \includegraphics[width=0.85\columnwidth]{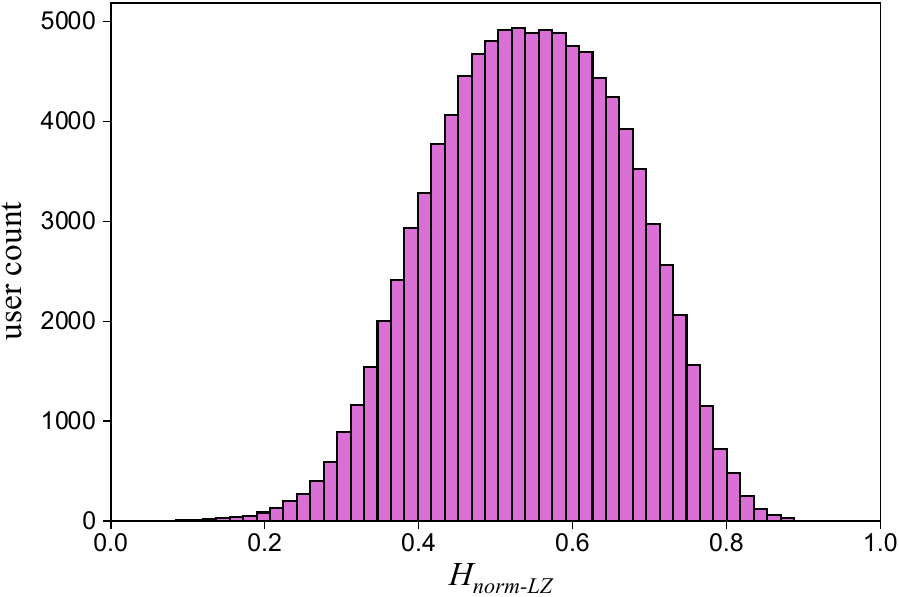}
    \caption{The distribution of $H_{\textit{norm-LZ}}$ for trajectories in \texttt{YJMob100K}.}
    \label{fig:entropy}
\end{figure}

Fig. \ref{fig:entropy} presents the distribution of $H_{\textit{norm-LZ}}$ for trajectories in the \texttt{YJMob100K} dataset, exhibiting a near-Gaussian distribution across the user population. Approximately 70\% of users exhibit entropy values between 0.4 and 0.65, indicating moderate levels of predictability that balance routine patterns with exploratory behaviors. Fewer than 2\% of users demonstrate extremely low entropy (below 0.2) or extremely high entropy (above 0.8), corresponding to highly repetitive or highly random movement patterns, respectively. Based on these statistics, we artificially divide the curriculum pretraining into three stages using augmented trajectories with increasing $H_{\textit{norm-LZ}}$ and \textit{Pho}, as given below:
\begin{itemize}
    \item Stage 1: $H_{\textit{norm-LZ}}<0.4$, \textit{Pho} is 3 days.
    \item Stage 2: $H_{\textit{norm-LZ}}<0.65$, \textit{Pho} is 7 days.
    \item Stage 3: all data with the complete \textit{Pho} of 15 days.
\end{itemize}

\subsubsection{MTL formulation}

The spatial area of the \texttt{YJMob100K} dataset is discretized into a grid, rendering the primary location estimation a classification task. For auxiliary distance and direction estimation, we also formulate these as classification tasks. Specifically, for distance estimation, the Euclidean distance \textit{dist} between consecutive trajectory points is discretized into four classes: stationary ($\textit{dist} < 500\ m$), short-range ($500\ m \le \textit{dist} < 1000\ m$), medium-range ($1000\ m \le \textit{dist} < 3500\ m$), and long-range ($\textit{dist} \ge 3500\ m$), corresponding to typical movement scales such as stationary, walking, running/cycling, and driving behaviors. Movement direction prediction discretizes the geographic direction into nine classes, including four cardinal directions (\textit{N}, \textit{E}, \textit{S}, \textit{W}), four ordinal directions (\textit{NE}, \textit{SE}, \textit{SW}, \textit{NW}), and stationary.

\section{Evaluation}

We evaluate our methods elaborated in Section \ref{sec:methodology} using the \texttt{YJMob100K} dataset introduced in Section \ref{sec:yjmob} with corresponding adaptation strategies described in Section \ref{sec:adaptation}.

\subsection{Experimental settings}

\begin{figure}[t]
    \centering
    \includegraphics[width=0.95\columnwidth]{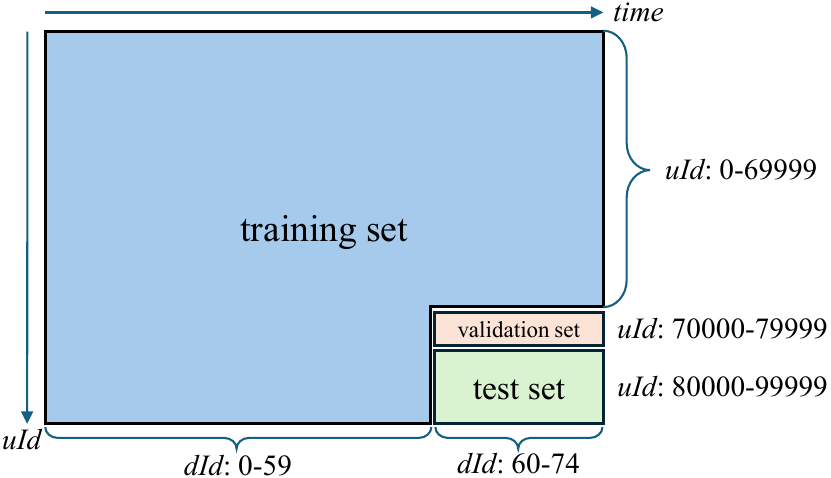}
    \caption{Training, validation, and test set split of the \texttt{YJMob100K} dataset.}
    \label{fig:split}
\end{figure}

The \texttt{YJMob100K} dataset is split by \textsl{uId} into training, validation, and test sets with a 7:1:2 ratio, as illustrated in Fig. \ref{fig:split}. We keep the test set identical to the \textit{HuMob challenge} for fair comparison. This split prevents data leakage and enables evaluation of the generalization capabilities of models for unseen users. The training set contains 70,000 users, the validation set 10,000 users, and the test set 20,000 users. On the training set, the model is trained on the first 60 days of user trajectories and tested on predicting the subsequent 15-day period (days 60-74). The same metrics used in the \textit{HuMob challenge} are adopted, namely GEO-BLEU \cite{10.1145/3557915.3560951} and DTW \cite{1104847}. GEO-BLEU measures the spatial coverage similarity between predicted and ground truth trajectories by computing \textit{n}-gram overlap with equal weighting and a brevity penalty parameter, providing insights into the model's ability to capture key spatial regions and movement hotspots. DTW evaluates the global spatiotemporal similarity between trajectory sequences by computing the minimum cumulative distance across all possible trajectory alignments using dynamic programming, effectively capturing both spatial accuracy and temporal sequence consistency while accommodating variable-length trajectories and temporal misalignments. Please refer to the cited papers for further details regarding these two metrics.

For experiments following the proposed curriculum learning pipeline, the model is pretrained for 60 epochs (20 for each stage) with an initial learning rate of $5\times10^{-5}$, followed by finetuning on real trajectories with a reduced initial learning rate of $2\times10^{-5}$. For experiments not following curriculum learning, the initial learning rate is also set to $5\times10^{-5}$. The Adam optimizer \cite{kingma2014adam} with a batch size of 128 is employed universally for all experiments. MoBERT employs 8 BERT encoder layers and 8 self-attention heads (also for the feature interaction module) with 256-dimensional latent spaces.

\subsection{Results}\label{sec:results}

Beginning with a baseline MoBERT architecture using only essential input features including $l=(x,y)$ and $t=\{\textsl{dId}, \textsl{tId}\}$, we gradually add
\begin{enumerate}
    \item \underline{S}emantics $s$ with $k=3$ or $7$ for \textsl{topKPOI}, obtaining MoBERT$_{\texttt{s3}}$ and MoBERT$_{\texttt{s7}}$;
    \item \underline{F}eature embedding interaction module (Section \ref{sec:interaction}), obtaining MoBERT$_{\texttt{s3/F}}$ (taking \texttt{s3} as an example);
    \item \underline{M}ulti-task learning (Section \ref{sec:model} and Fig. \ref{fig:multitask}), obtaining MoBERT$_{\texttt{s3/F/M}}$;
    \item \underline{E}ntropy-driven curriculum learning (Section \ref{sec:entropy2} and \ref{sec:curriculum-stages}), obtaining MoBERT$_{\texttt{s3/F/M/E}}$;
\end{enumerate}
to observe the contribution of each components and naturally form ablation studies. We also compare the performance of the top three in the \textit{HuMob Challenge} 2023, namely LP-BERT (\textit{1st}) \cite{10.1145/3615894.3628498}, GeoFormer (\textit{2nd}) \cite{10.1145/3615894.3628499}, MOBB (\textit{3rd}) \cite{10.1145/3615894.3628500}, where the leaderboard is online here\footnote{\url{https://connection.mit.edu/humob-challenge-2023/}} (last accessed: Aug. 28, 2025).

Table \ref{tab:main} presents the performance of MoBERT variants and the top three methods in the \textit{HuMob Challenge} 2023, as evaluated by GEO-BLEU and DTW. It can be observed that MoBERT$_{\texttt{s3/F/M/E}}$ with all four optimizations achieves state-of-the-art performance on both metrics, demonstrating the effectiveness of the proposed approaches. Particularly for DTW, which quantifies the temporal and spatial alignment of entire trajectories, MoBERT$_{\texttt{s3/F/M/E}}$ achieves 12.7\% lower (better) DTW values than LP-BERT, demonstrating that MoBERT$_{\texttt{s3/F/M/E}}$ can capture both global trends and local variations more accurately than competing methods.

We highlight the performance gains of current MoBERT variants in parentheses compared to their previous variants to demonstrate the contribution of newly applied optimization strategies. Note that both MoBERT$_{\texttt{s3}}$ and MoBERT$_{\texttt{s7}}$ are compared with the vanilla MoBERT. These results reveal that MTL contributes most significantly to GEO-BLEU, while entropy-driven curriculum learning optimizes DTW most effectively. MoBERT$_{\texttt{s3}}$ and MoBERT$_{\texttt{s7}}$ achieve superior GEO-BLEU and DTW performance, respectively. We select MoBERT$_{\texttt{s3}}$ for the following experiments as the \textit{HuMob Challenge} prioritizes GEO-BLEU over DTW. Furthermore, all four optimization strategies improve performance on both metrics compared to cases when they are absent, demonstrating that all four strategies are effective for human mobility prediction.

Table \ref{tab:speedup} compares the convergence speed by counting epochs required to reach target training/validation losses between MoBERT$_{\texttt{s3/F/M/E}}$ and MoBERT$_{\texttt{s3/F/M}}$ (with and without curriculum learning). Although the contribution of entropy-driven curriculum learning to GEO-BLEU is not the largest, it dramatically accelerates the training process and achieves up to 2.92-fold convergence speed, which supports our analysis of curriculum learning and validates that the proposed $H_{\textit{norm-LZ}}$ is an appropriate difficulty measurement for human mobility data. Our final model corresponds to a validation loss close to 2.1, so the achieved acceleration ratio has considerable reference value.

\begin{table}[t]
\renewcommand{\arraystretch}{1.2}
\centering
\caption{Performance comparison of MoBERT variants and other studies on GEO-BLEU (larger is better) and DTW (smaller is better). Numbers in parentheses show differences between the current MoBERT variant and the previous MoBERT variant.}
\begin{tabular}{@{}rcc@{}}
\toprule
\textbf{Model} & GEO-BLEU ($\uparrow$) & DTW ($\downarrow$) \\ 
\midrule
MoBERT$_{\texttt{\phantom{s3/F/M/E}}}$   & 0.264                   & 30.06 \\
MoBERT$_{\texttt{s3\phantom{/F/M/E}}}$   & 0.287 (+0.023)          & 29.11 (-0.95) \\
MoBERT$_{\texttt{s7\phantom{/F/M/E}}}$   & 0.284 (+0.020)          & 28.98 (-1.08) \\
MoBERT$_{\texttt{s3/F\phantom{/M/E}}}$   & 0.307 (+0.020)          & 28.93 (-0.18) \\
MoBERT$_{\texttt{s3/F/M\phantom{/E}}}$   & 0.335 (\textbf{+0.028}) & 28.16 (-0.77) \\
MoBERT$_{\texttt{s3/F/M/E}}$             & \textbf{0.354} (+0.019) & \textbf{26.15} (\textbf{-2.01}) \\
MOBB (3$^{\textit{rd}}$) \cite{10.1145/3615894.3628500}      & 0.327 & 38.65 \\
GeoFormer (2$^{\textit{nd}}$) \cite{10.1145/3615894.3628499} & 0.316 & 26.22 \\
LP-BERT (1$^{\textit{st}}$) \cite{10.1145/3615894.3628498}   & 0.344 & 29.96 \\
\bottomrule
\end{tabular}
\label{tab:main}
\end{table}

\begin{table}[t]
\renewcommand{\arraystretch}{1.2}
\centering
\caption{Epochs required to reach target losses for MoBERT$_{\texttt{s3/F/M/E}}$ and MoBERT$_{\texttt{s3/F/M}}$ (w/ or w/o curriculum learning).}
\begin{tabular}{@{}rcc@{}}
\toprule
Target losses & MoBERT$_{\texttt{s3/F/M/E}}$ & MoBERT$_{\texttt{s3/F/M}}$ \\ 
\midrule
training loss $\le 2.6$   & 37 (1.95-fold speed) & 72   \\
validation loss $\le 2.1$ & 38 (2.92-fold speed) & 111  \\
\bottomrule
\end{tabular}
\label{tab:speedup}
\end{table}

\subsection{Hyperparameter selection}

We now adopt grid search for hyperparameter selection. Table \ref{tab:mtl} lists the MTL results with varying $\lambda_1$ and $\lambda_2$. Starting with MoBERT$_{\texttt{s3/F}}$, auxiliary tasks are introduced with different weights. Introducing only the direction prediction task and gradually increasing its weight from 0.2 to 0.8 results in a 5.6\% improvement in GEO-BLEU and a 2.1\% decrease in DTW. However, further increasing the weight to 1.0 leads to a decline in model performance. This suggests that moderately increasing the direction task weight enables the model to more accurately capture trajectory spatial trends, while excessively high weights may undermine the learning of the primary task. A similar pattern is observed for the movement distance prediction task: when the distance estimation task weight is set to 0.5, the model achieves the highest GEO-BLEU value. The largest improvement arises when both sub-tasks are active, where $\lambda_1$ and $\lambda_2$ are set to 0.5 and 0.8, respectively, indicating that two sub-tasks provide complementary constraints and lead to more stable and accurate estimation. We select the best weights for single tasks and identify the optimal pair of weights, where $\lambda_1$ and $\lambda_2$ are set to 0.5 and 0.8, respectively. These settings are maintained for MoBERT$_{\texttt{s3/F/M}}$ and MoBERT$_{\texttt{s3/F/M/E}}$, with their performances presented in Table \ref{tab:main} and \ref{tab:speedup}.

We also determine the optimal model hyperparameters, including the number of encoder layers and the number of self-attention heads. Table \ref{tab:hyper} illustrates the results of MoBERT$_{\texttt{s3/F/M/E}}$ with varying model hyperparameters. A larger number of encoder layers yields improved Geo-BLEU and DTW scores. We do not evaluate additional encoder layers due to resource limitations. By contrast, increasing the number of self-attention heads produces lower accuracy. More heads allocate each head a smaller sub-dimension, thereby reducing its representation capacity. Therefore, we select 8 encoder layers and 8 self-attention heads for model construction.

\begin{table}[t]
\renewcommand{\arraystretch}{1.2}
\centering
\caption{Grid search results of MTL with varying $\lambda_1$ and $\lambda_2$.}
\begin{tabular}{@{}rccc@{}}
\toprule
auxiliary task & weight & GEO-BLEU ($\uparrow$) & DTW ($\downarrow$) \\ 
\midrule
none (MoBERT$_{\texttt{s3/F}}$) & - & 0.307 & 28.93 \\
\midrule
distance estimation & $\lambda_1=0.2$ & 0.323 & 29.15 \\
distance estimation & $\lambda_1=0.5$ & \textbf{0.324} & \textbf{28.87} \\
distance estimation & $\lambda_1=0.8$ & 0.322 & 28.77 \\
distance estimation & $\lambda_1=1.0$ & 0.321 & 29.02 \\
\midrule
direction estimation & $\lambda_2=0.2$ & 0.320 & 28.78 \\
direction estimation & $\lambda_2=0.5$ & 0.322 & 28.47 \\
direction estimation & $\lambda_2=0.8$ & \textbf{0.325} & \textbf{28.32} \\
direction estimation & $\lambda_2=1.0$ & 0.321 & 28.52 \\
\midrule
both (MoBERT$_{\texttt{s3/F/M}}$) & \begin{tabular}[c]{@{}l@{}}$\lambda_1=0.5$\\$\lambda_2=0.8$\end{tabular} & \textbf{0.335} & \textbf{28.16} \\ 
\bottomrule
\end{tabular}
\label{tab:mtl}
\end{table}

\begin{table}[t]
\renewcommand{\arraystretch}{1.2}
\centering
\caption{Results of MoBERT$_{\texttt{s3/F/M/E}}$ with different model hyperparameters.}
\begin{tabular}{@{}cccc@{}}
\toprule
\# encoder layers & \# heads & GEO-BLEU ($\uparrow$) & DTW ($\downarrow$) \\ 
\midrule
8 & 8  & \textbf{0.354} & \textbf{26.15} \\
8 & 16 & 0.309 & 35.16 \\
4 & 8  & 0.343 & 26.83 \\
4 & 16 & 0.326 & 31.19 \\
\bottomrule
\end{tabular}
\label{tab:hyper}
\end{table}

\subsection{Cross-city generalization}

\begin{figure}[t]
    \centering
    \includegraphics[width=1.00\columnwidth]{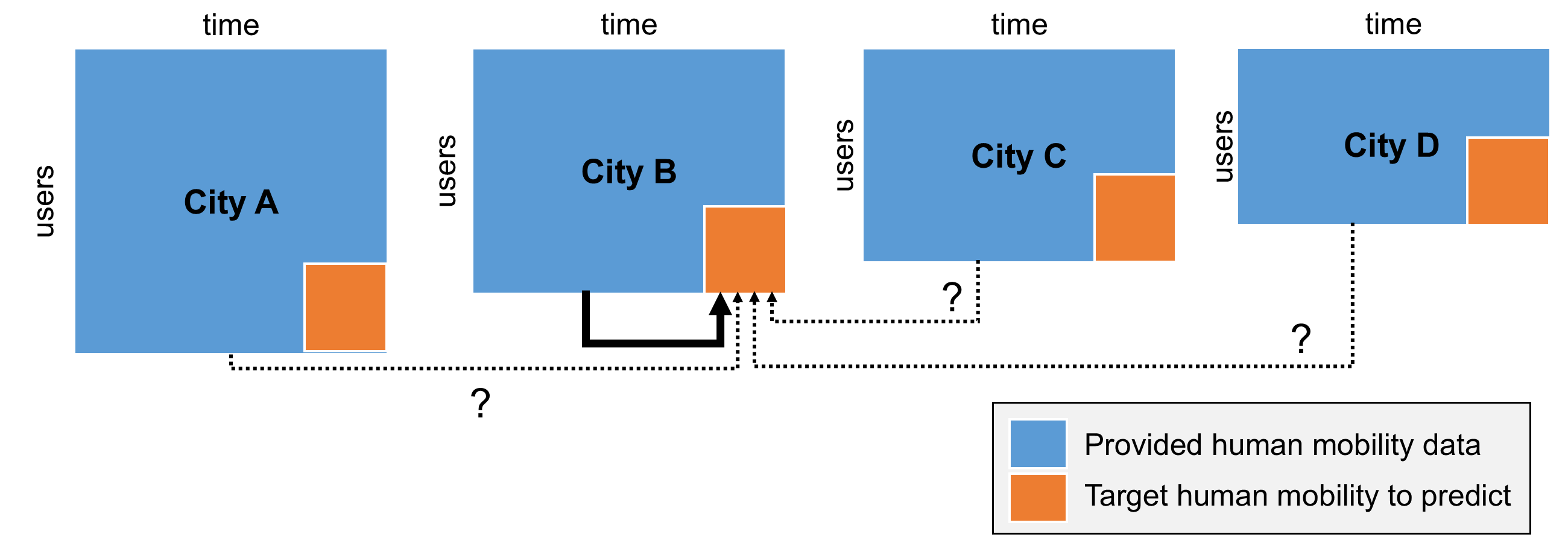}
    \caption{\textit{Multi-City Human Mobility Prediction} problem setting in the \textit{HuMob Challenge} 2024 (adapted from \cite{10.1145/3681771}).}
    \label{fig:crosscity}
\end{figure}

\begin{table}[t]
\renewcommand{\arraystretch}{1.2}
\centering
\caption{City-wise performance comparison.}
\begin{tabular}{@{}cccc@{}}
\toprule
Model & Data & GEO-BLEU ($\uparrow$) & DTW ($\downarrow$) \\ 
\midrule
Llama-3-8B-Mob \cite{10.1145/3681771.3699908} & City B & \textbf{0.354} & 25.39 \\
1st place in 2024                             & City C & 0.296 & \textbf{20.57} \\
(trained on City A+B)                         & City D & 0.321 & 31.94 \\
\midrule
LP-BERT \cite{10.1145/3615894.3628498}& City B & 0.309 & \textbf{23.30} \\
1st place in 2023                     & City C & 0.268 & 23.81 \\
(trained on all cities)               & City D & 0.303 & 38.89 \\
\midrule
MoBERT$_{\texttt{s3/F/M/E}}$ & City B & 0.329 & 23.43 \\
Ours                         & City C & \textbf{0.314} & \textbf{18.33} \\
(trained on City A only)     & City D & \textbf{0.328} & 38.03 \\
\bottomrule
\end{tabular}
\label{tab:cities}
\end{table}

Human mobility models often face different geographic settings, transport infrastructure, and behavioral patterns across cities. A model trained in a single city may therefore transfer poorly. The \textit{HuMob Challenge} 2024 highlighted this question under the theme \textit{Multi-City Human Mobility Prediction}, and asked whether data from additional cities can boost the target-city forecast, as depicted in Fig. \ref{fig:crosscity}. The dataset is enlarged to four cites, denoted as City A, B, C, and D, where City A is the dataset used for all previous evaluations (as shown in Fig. \ref{fig:split}). For each city, 3,000 user trajectories formed the test set. 

We directly use MoBERT$_{\texttt{s3/F/M/E}}$ trained on City A to perform inference on the other three cities without any fine-tuning. To ensure fairness, the test sets are exactly the same as those used in the \textit{HuMob Challenge} 2024. Table \ref{tab:cities} presents a city-wise performance comparison with Llama-3-8B-Mob (\textit{1st} in 2024) \cite{10.1145/3681771.3699908} and LP-BERT (\textit{1st} in 2023) \cite{10.1145/3615894.3628498}. MoBERT$_{\texttt{s3/F/M/E}}$ achieves an average Geo-BLEU of 0.324, matching the performance of Llama-3-8B-Mob while substantially outperforming LP-BERT. On City B, where Llama-3-8B-Mob benefits from training exposure, MoBERT$_{\texttt{s3/F/M/E}}$ performs within 7\% of the LLM-based model while significantly outperforming LP-BERT despite its access to City B. More notably, MoBERT$_{\texttt{s3/F/M/E}}$ achieves the highest Geo-BLEU scores on both City C and D, surpassing both competitors. Similar patterns also emerge in DTW metrics, particularly for City C, indicating robust trajectory shape fitting capabilities for unseen urban environments.

These results reveal that architectural design and training strategies can be more impactful than simply expanding training data coverage across multiple cities. The zero-shot generalization capability of MoBERT$_{\texttt{s3/F/M/E}}$ with only 7.02 million parameters, approximately 1/6 the size of Llama-3-8B-Mob, demonstrates that compact models with well-designed entropy-driven curriculum learning and MTL can efficiently extract transferable mobility patterns. Conversely, the underwhelming performance of LP-BERT despite access to the most comprehensive multi-city training data suggests that naive data aggregation may introduce conflicting patterns that impede effective learning. The findings indicate that model capacity for extracting generalizable features, rather than geographic data diversity alone, determines cross-city transferability, with the proposed training approaches, potentially enabling adaptation to diverse urban contexts without finetuning.

\section{Conclusion}

This paper presents a unified training framework that integrates entropy-driven curriculum learning with MTL for human mobility prediction. The entropy-driven curriculum learning strategy employs normalized LZ compression to quantify trajectory predictability, organizing training progression from simple to complex patterns based on information-theoretic principles grounded in Fano's inequality. This approach achieves up to 2.92-fold convergence speed compared to conventional training by random sampling.

The MTL component simultaneously optimizes location prediction alongside auxiliary distance and direction estimation tasks, providing complementary supervision signals that are universally available in any mobility dataset. The proposed MoBERT architecture incorporates multi-feature embeddings and attention-based feature interaction mechanisms to capture complex spatiotemporal dependencies.

Experimental results on the \texttt{YJMob100K} dataset demonstrate state-of-the-art performance, achieving a GEO-BLEU score of 0.354 and a DTW distance of 26.15, outperforming existing methods including top performers from the \textit{HuMob Challenge} 2023. Cross-city generalization experiments reveal robust transferability, with the method achieving competitive performance on unseen urban environments while using significantly fewer parameters than other approaches. The method trained exclusively on one city achieves comparable or superior results to models trained on multiple cities, indicating effective extraction of generalizable mobility patterns.

\section*{Acknowledgment}

We would like to thank all members of the Professorship of Big Geospatial Data Management at Technical University of Munich for their insightful discussions and comments.

\bibliographystyle{IEEEtran}
\bibliography{IEEEabrv,reference}

@inproceedings{devlin-etal-2019-bert,
  title = "{BERT}: Pre-training of Deep Bidirectional Transformers for Language Understanding",
  author = "Devlin, Jacob  and
      Chang, Ming-Wei  and
      Lee, Kenton  and
      Toutanova, Kristina",
  editor = "Burstein, Jill  and
      Doran, Christy  and
      Solorio, Thamar",
  booktitle = "Proceedings of the 2019 Conference of the North {A}merican Chapter of the Association for Computational Linguistics: Human Language Technologies, Volume 1 (Long and Short Papers)",
  month = jun,
  year = "2019",
  address = "Minneapolis, Minnesota",
  publisher = "Association for Computational Linguistics",
  doi = "10.18653/v1/N19-1423",
  pages = "4171--4186",
  url = "https://aclanthology.org/N19-1423/",
}

@inproceedings{NIPS2017_3f5ee243,
  author = {Vaswani, Ashish and Shazeer, Noam and Parmar, Niki and Uszkoreit, Jakob and Jones, Llion and Gomez, Aidan N and Kaiser, \L ukasz and Polosukhin, Illia},
  booktitle = {Advances in Neural Information Processing Systems},
  editor = {I. Guyon and U. Von Luxburg and S. Bengio and H. Wallach and R. Fergus and S. Vishwanathan and R. Garnett},
  pages = {},
  publisher = {Curran Associates, Inc.},
  title = {Attention is All you Need},
  volume = {30},
  year = {2017},
  url = {https://papers.neurips.cc/paper/7181-attention-is-all-you-need.pdf}
}

@inproceedings{10.1145/3557915.3560951,
  author = {Shimizu, Toru and Tsubouchi, Kota and Yabe, Takahiro},
  title = {{GEO-BLEU}: similarity measure for geospatial sequences},
  year = {2022},
  isbn = {9781450395298},
  publisher = {Association for Computing Machinery},
  address = {New York, NY, USA},
  url = {https://doi.org/10.1145/3557915.3560951},
  doi = {10.1145/3557915.3560951},
  booktitle = {Proceedings of the 30th International Conference on Advances in Geographic Information Systems},
  articleno = {17},
  numpages = {4},
  location = {Seattle, Washington},
  series = {SIGSPATIAL '22}
}

@article{song2010limits,
	author = {Song, Chaoming and Qu, Zehui and Blumm, Nicholas and Barab{\'a}si, Albert-L{\'a}szl{\'o}},
	journal = {Science},
	number = {5968},
	year = {2010},
	pages = {1018--1021},
	publisher = {American Association for the Advancement of Science (AAAS)},
	title = {Limits of {Predictability} in {Human} {Mobility}},
	volume = {327},
}

@INPROCEEDINGS{7840811,
  author={Zhao, Kai and Tarkoma, Sasu and Liu, Siyuan and Vo, Huy},
  booktitle={2016 IEEE International Conference on Big Data (Big Data)}, 
  title={Urban human mobility data mining: An overview}, 
  year={2016},
  volume={},
  number={},
  pages={1911-1920},
  doi={10.1109/BigData.2016.7840811}
}

@article{10.1177/23998083221075634,
  author = {Masahiko Haraguchi and Akihiko Nishino and Akira Kodaka and Maura Allaire and Upmanu Lall and Liao Kuei-Hsien and Kaya Onda and Kota Tsubouchi and Naohiko Kohtake},
  title ={Human mobility data and analysis for urban resilience: A systematic review},
  journal = {Environment and Planning B: Urban Analytics and City Science},
  volume = {49},
  number = {5},
  pages = {1507-1535},
  year = {2022},
  doi = {10.1177/23998083221075634},
  URL = {https://doi.org/10.1177/23998083221075634},
  eprint = {https://doi.org/10.1177/23998083221075634}
}

@inproceedings{10.5555/3060832.3060987,
  author = {Song, Xuan and Kanasugi, Hiroshi and Shibasaki, Ryosuke},
  title = {Deeptransport: prediction and simulation of human mobility and transportation mode at a citywide level},
  year = {2016},
  isbn = {9781577357704},
  publisher = {AAAI Press},
  booktitle = {Proceedings of the Twenty-Fifth International Joint Conference on Artificial Intelligence},
  pages = {2618–2624},
  numpages = {7},
  location = {New York, New York, USA},
  series = {IJCAI'16}
}

@Article{Pappalardo2023,
  author={Pappalardo, Luca and Manley, Ed and Sekara, Vedran and Alessandretti, Laura},
  title={Future directions in human mobility science},
  journal={Nature Computational Science},
  year={2023},
  month={Jul},
  day={01},
  volume={3},
  number={7},
  pages={588-600},
  issn={2662-8457},
  doi={10.1038/s43588-023-00469-4},
  url={https://doi.org/10.1038/s43588-023-00469-4}
}

@INPROCEEDINGS{7373330,
  author={Lian, Defu and Ge, Yong and Zhang, Fuzheng and Yuan, Nicholas Jing and Xie, Xing and Zhou, Tao and Rui, Yong},
  booktitle={2015 IEEE International Conference on Data Mining}, 
  title={Content-Aware Collaborative Filtering for Location Recommendation Based on Human Mobility Data}, 
  year={2015},
  volume={},
  number={},
  pages={261-270},
  doi={10.1109/ICDM.2015.69}
}

@INPROCEEDINGS{9861127,
  author={Alix, Gian and Yanin, Nina and Pechlivanoglou, Tilemachos and Li, Jing and Heidari, Farzaneh and Papagelis, Manos},
  booktitle={2022 23rd IEEE International Conference on Mobile Data Management (MDM)}, 
  title={A Mobility-based Recommendation System for Mitigating the Risk of Infection during Epidemics}, 
  year={2022},
  volume={},
  number={},
  pages={292-295},
  doi={10.1109/MDM55031.2022.00063}
}

@article{Jiang_Song_Fan_2018,
  title={{DeepUrbanMomentum}: An Online Deep-Learning System for Short-Term Urban Mobility Prediction},
  volume={32},
  url={https://ojs.aaai.org/index.php/AAAI/article/view/11338},
  DOI={10.1609/aaai.v32i1.11338},
  number={1},
  journal={Proceedings of the AAAI Conference on Artificial Intelligence},
  author={Jiang, Renhe and Song, Xuan and Fan, Zipei and Xia, Tianqi and Chen, Quanjun and Miyazawa, Satoshi and Shibasaki, Ryosuke},
  year={2018},
  month={Apr.}
}

@article{Liu_Wu_Wang_2016,
  title={Predicting the Next Location: A Recurrent Model with Spatial and Temporal Contexts},
  volume={30},
  url={https://ojs.aaai.org/index.php/AAAI/article/view/9971},
  DOI={10.1609/aaai.v30i1.9971},
  number={1},
  journal={Proceedings of the AAAI Conference on Artificial Intelligence},
  author={Liu, Qiang and Wu, Shu and Wang, Liang and Tan, Tieniu},
  year={2016},
  month={Feb.}
}

@inproceedings{ijcai2018p324,
  title     = {{HST-LSTM}: A Hierarchical Spatial-Temporal Long-Short Term Memory Network for Location Prediction},
  author    = {Dejiang Kong and Fei Wu},
  booktitle = {Proceedings of the Twenty-Seventh International Joint Conference on
               Artificial Intelligence, {IJCAI-18}},
  publisher = {International Joint Conferences on Artificial Intelligence Organization},
  pages     = {2341--2347},
  year      = {2018},
  month     = {7},
  doi       = {10.24963/ijcai.2018/324},
  url       = {https://doi.org/10.24963/ijcai.2018/324},
}

@article{WANG2024129872,
  title = {Coupling graph neural networks and travel mode choice for human mobility prediction},
  journal = {Physica A: Statistical Mechanics and its Applications},
  volume = {646},
  pages = {129872},
  year = {2024},
  issn = {0378-4371},
  doi = {https://doi.org/10.1016/j.physa.2024.129872},
  url = {https://www.sciencedirect.com/science/article/pii/S0378437124003819},
  author = {Kun Wang and Zhenghong Peng and Meng Cai and Hao Wu and Lingbo Liu and Zhihao Sun},
}

@article{Liu_Rong_Guo_2023,
  title={Human Mobility Modeling during the {COVID-19} Pandemic via Deep Graph Diffusion Infomax},
  volume={37},
  url={https://ojs.aaai.org/index.php/AAAI/article/view/26678},
  DOI={10.1609/aaai.v37i12.26678},
  number={12},
  journal={Proceedings of the AAAI Conference on Artificial Intelligence},
  author={Liu, Yang and Rong, Yu and Guo, Zhuoning and Chen, Nuo and Xu, Tingyang and Tsung, Fugee and Li, Jia},
  year={2023},
  month={Jun.},
  pages={14347-14355}
}

@inproceedings{10.1145/3178876.3186058,
  author = {Feng, Jie and Li, Yong and Zhang, Chao and Sun, Funing and Meng, Fanchao and Guo, Ang and Jin, Depeng},
  title = {{DeepMove}: Predicting Human Mobility with Attentional Recurrent Networks},
  year = {2018},
  isbn = {9781450356398},
  publisher = {International World Wide Web Conferences Steering Committee},
  address = {Republic and Canton of Geneva, CHE},
  url = {https://doi.org/10.1145/3178876.3186058},
  doi = {10.1145/3178876.3186058},
  booktitle = {Proceedings of the 2018 World Wide Web Conference},
  pages = {1459–1468},
  numpages = {10},
  location = {Lyon, France},
  series = {WWW '18}
}

@article{10.1145/2501654.2501656,
  author = {Parent, Christine and Spaccapietra, Stefano and Renso, Chiara and Andrienko, Gennady and Andrienko, Natalia and Bogorny, Vania and Damiani, Maria Luisa and Gkoulalas-Divanis, Aris and Macedo, Jose and Pelekis, Nikos and Theodoridis, Yannis and Yan, Zhixian},
title = {Semantic trajectories modeling and analysis},
  year = {2013},
  issue_date = {August 2013},
  publisher = {Association for Computing Machinery},
  address = {New York, NY, USA},
  volume = {45},
  number = {4},
  issn = {0360-0300},
  url = {https://doi.org/10.1145/2501654.2501656},
  doi = {10.1145/2501654.2501656},
  journal = {ACM Comput. Surv.},
  month = aug,
  articleno = {42},
  numpages = {32}
}

@article{10.1145/2483669.2483682,
  author = {Yan, Zhixian and Chakraborty, Dipanjan and Parent, Christine and Spaccapietra, Stefano and Aberer, Karl},
  title = {Semantic trajectories: Mobility data computation and annotation},
  year = {2013},
  issue_date = {June 2013},
  publisher = {Association for Computing Machinery},
  address = {New York, NY, USA},
  volume = {4},
  number = {3},
  issn = {2157-6904},
  url = {https://doi.org/10.1145/2483669.2483682},
  doi = {10.1145/2483669.2483682},
  journal = {ACM Trans. Intell. Syst. Technol.},
  month = jul,
  articleno = {49},
  numpages = {38},
}

@ARTICLE{9430503,
  author={Wang, Huandong and Li, Yong and Jin, Depeng and Han, Zhu},
  journal={IEEE Journal on Selected Areas in Communications}, 
  title={Attentional Markov Model for Human Mobility Prediction}, 
  year={2021},
  volume={39},
  number={7},
  pages={2213-2225},
  doi={10.1109/JSAC.2021.3078499}
}

@inproceedings{10.1145/3615894.3628501,
  author = {Suzuki, Masahiro and Furuta, Shomu and Fukazawa, Yusuke},
  title = {Personalized human mobility prediction for HuMob challenge},
  year = {2023},
  isbn = {9798400703560},
  publisher = {Association for Computing Machinery},
  address = {New York, NY, USA},
  url = {https://doi.org/10.1145/3615894.3628501},
  doi = {10.1145/3615894.3628501},
  booktitle = {Proceedings of the 1st International Workshop on the Human Mobility Prediction Challenge},
  pages = {22–25},
  numpages = {4},
  location = {Hamburg, Germany},
  series = {HuMob-Challenge '23}
}

@ARTICLE{8812927,
  author={Shi, Hongzhi and Li, Yong and Cao, Hancheng and Zhou, Xiangxin and Zhang, Chao and Kostakos, Vassilis},
  journal={IEEE Transactions on Knowledge and Data Engineering}, 
  title={Semantics-Aware Hidden Markov Model for Human Mobility}, 
  year={2021},
  volume={33},
  number={3},
  pages={1183-1194},
  doi={10.1109/TKDE.2019.2937296}
}

@article{QIAO201899,
  title = {A hybrid Markov-based model for human mobility prediction},
  journal = {Neurocomputing},
  volume = {278},
  pages = {99-109},
  year = {2018},
  note = {Recent Advances in Machine Learning for Non-Gaussian Data Processing},
  issn = {0925-2312},
  doi = {https://doi.org/10.1016/j.neucom.2017.05.101},
  url = {https://www.sciencedirect.com/science/article/pii/S0925231217314455},
  author = {Yuanyuan Qiao and Zhongwei Si and Yanting Zhang and Fehmi Ben Abdesslem and Xinyu Zhang and Jie Yang},
}

@ARTICLE{9392296,
  author={Wang, Xin and Chen, Yudong and Zhu, Wenwu},
  journal={IEEE Transactions on Pattern Analysis and Machine Intelligence}, 
  title={A Survey on Curriculum Learning}, 
  year={2022},
  volume={44},
  number={9},
  pages={4555-4576},
  doi={10.1109/TPAMI.2021.3069908}
}

@inproceedings{10.1145/1553374.1553380,
  author = {Bengio, Yoshua and Louradour, J\'{e}r\^{o}me and Collobert, Ronan and Weston, Jason},
  title = {Curriculum learning},
  year = {2009},
  isbn = {9781605585161},
  publisher = {Association for Computing Machinery},
  address = {New York, NY, USA},
  url = {https://doi.org/10.1145/1553374.1553380},
  doi = {10.1145/1553374.1553380},
  booktitle = {Proceedings of the 26th Annual International Conference on Machine Learning},
  pages = {41–48},
  numpages = {8},
  location = {Montreal, Quebec, Canada},
  series = {ICML '09}
}

@inproceedings{NIPS2010_e57c6b95,
 author = {Kumar, M. and Packer, Benjamin and Koller, Daphne},
 booktitle = {Advances in Neural Information Processing Systems},
 editor = {J. Lafferty and C. Williams and J. Shawe-Taylor and R. Zemel and A. Culotta},
 pages = {},
 publisher = {Curran Associates, Inc.},
 title = {Self-Paced Learning for Latent Variable Models},
 volume = {23},
 year = {2010}
}

@Article{Yabe2024,
  author={Yabe, Takahiro and Tsubouchi, Kota and Shimizu, Toru and Sekimoto, Yoshihide and Sezaki, Kaoru and Moro, Esteban and Pentland, Alex},
  title={YJMob100K: City-scale and longitudinal dataset of anonymized human mobility trajectories},
  journal={Scientific Data},
  year={2024},
  month={Apr},
  day={18},
  volume={11},
  number={1},
  pages={397},
  issn={2052-4463},
  doi={10.1038/s41597-024-03237-9},
  url={https://doi.org/10.1038/s41597-024-03237-9}
}

@article{LEI2025102597,
  title = {A deep multimodal network for multi-task trajectory prediction},
  journal = {Information Fusion},
  volume = {113},
  pages = {102597},
  year = {2025},
  issn = {1566-2535},
  doi = {10.1016/j.inffus.2024.102597},
  author = {Da Lei and Min Xu and Shuaian Wang},
}

@ARTICLE{9130086,
  author={Feng, Jie and Li, Yong and Yang, Zeyu and Qiu, Qiang and Jin, Depeng},
  journal={IEEE Transactions on Knowledge and Data Engineering}, 
  title={Predicting Human Mobility With Semantic Motivation via Multi-Task Attentional Recurrent Networks}, 
  year={2022},
  volume={34},
  number={5},
  pages={2360-2374},
  doi={10.1109/TKDE.2020.3006048}
}

@Article{Caruana1997,
  author={Caruana, Rich},
  title={Multitask Learning},
  journal={Machine Learning},
  year={1997},
  month={Jul},
  day={01},
  volume={28},
  number={1},
  pages={41-75},
  issn={1573-0565},
  doi={10.1023/A:1007379606734},
  url={https://doi.org/10.1023/A:1007379606734}
}

@ARTICLE{1055672,
  author={Lempel, A. and Cohn, M. and Eastman, W.},
  journal={IEEE Transactions on Information Theory}, 
  title={A class of balanced binary sequences with optimal autocorrelation properties},
  year={1977},
  volume={23},
  number={1},
  pages={38-42},
  doi={10.1109/TIT.1977.1055672}
}

@proceedings{10.1145/3615894,
  title = {HuMob-Challenge '23: Proceedings of the 1st International Workshop on the Human Mobility Prediction Challenge},
  year = {2023},
  isbn = {9798400703560},
  publisher = {Association for Computing Machinery},
  address = {New York, NY, USA},
  location = {Hamburg, Germany}
}

@proceedings{10.1145/3681771,
  title = {HuMob'24: Proceedings of the 2nd ACM SIGSPATIAL International Workshop on Human Mobility Prediction Challenge},
  year = {2024},
  isbn = {9798400711503},
  publisher = {Association for Computing Machinery},
  address = {New York, NY, USA},
  location = {Atlanta, GA, USA}
}

@ARTICLE{1104847,
  author={Bellman, R. and Kalaba, R.},
  journal={IRE Transactions on Automatic Control}, 
  title={On adaptive control processes}, 
  year={1959},
  volume={4},
  number={2},
  pages={1-9},
  doi={10.1109/TAC.1959.1104847}
}

@ARTICLE{9204396,
  author={Wang, Senzhang and Cao, Jiannong and Yu, Philip S.},
  journal={IEEE Transactions on Knowledge and Data Engineering}, 
  title={Deep Learning for Spatio-Temporal Data Mining: A Survey}, 
  year={2022},
  volume={34},
  number={8},
  pages={3681-3700},
  doi={10.1109/TKDE.2020.3025580}
}

@INPROCEEDINGS{9378429,
  author={Fiorini, Stefano and Pilotti, Giorgio and Ciavotta, Michele and Maurino, Andrea},
  booktitle={2020 IEEE International Conference on Big Data (Big Data)}, 
  title={3D-CLoST: A CNN-LSTM Approach for Mobility Dynamics Prediction in Smart Cities}, 
  year={2020},
  volume={},
  number={},
  pages={3180-3189},
  doi={10.1109/BigData50022.2020.9378429}
}

@inproceedings{10.1145/3219819.3219931,
  author = {Shen, Bilong and Liang, Xiaodan and Ouyang, Yufeng and Liu, Miaofeng and Zheng, Weimin and Carley, Kathleen M.},
  title = {StepDeep: A Novel Spatial-temporal Mobility Event Prediction Framework based on Deep Neural Network},
  year = {2018},
  isbn = {9781450355520},
  publisher = {Association for Computing Machinery},
  address = {New York, NY, USA},
  url = {https://doi.org/10.1145/3219819.3219931},
  doi = {10.1145/3219819.3219931},
  booktitle = {Proceedings of the 24th ACM SIGKDD International Conference on Knowledge Discovery \& Data Mining},
  pages = {724–733},
  numpages = {10},
  location = {London, United Kingdom},
  series = {KDD '18}
}

@inproceedings{osovitskiy2021an,
  title={An Image is Worth 16x16 Words: Transformers for Image Recognition at Scale},
  author={Alexey Dosovitskiy and Lucas Beyer and Alexander Kolesnikov and Dirk Weissenborn and Xiaohua Zhai and Thomas Unterthiner and Mostafa Dehghani and Matthias Minderer and Georg Heigold and Sylvain Gelly and Jakob Uszkoreit and Neil Houlsby},
  booktitle={International Conference on Learning Representations},
  year={2021},
  url={https://openreview.net/forum?id=YicbFdNTTy}
}

@article{Wang_Chen_Pan_2023,
  title={Easy Begun Is Half Done: Spatial-Temporal Graph Modeling with ST-Curriculum Dropout},
  volume={37},
  url={https://ojs.aaai.org/index.php/AAAI/article/view/25590},
  DOI={10.1609/aaai.v37i4.25590},
  number={4},
  journal={Proceedings of the AAAI Conference on Artificial Intelligence},
  author={Wang, Hongjun and Chen, Jiyuan and Pan, Tong and Fan, Zipei and Song, Xuan and Jiang, Renhe and Zhang, Lingyu and Xie, Yi and Wang, Zhongyi and Zhang, Boyuan},
  year={2023},
  month={Jun.},
  pages={4668-4675}
}

@ARTICLE{335943,
  author={Te Sun Han and Verdu, S.},
  journal={IEEE Transactions on Information Theory}, 
  title={Generalizing the Fano inequality}, 
  year={1994},
  volume={40},
  number={4},
  pages={1247-1251},
  doi={10.1109/18.335943}
}

@ARTICLE{1055714,
  author={Ziv, J. and Lempel, A.},
  journal={IEEE Transactions on Information Theory}, 
  title={A universal algorithm for sequential data compression}, 
  year={1977},
  volume={23},
  number={3},
  pages={337-343},
  doi={10.1109/TIT.1977.1055714}
}

@article{kingma2014adam,
  title={Adam: A method for stochastic optimization},
  author={Kingma, Diederik P and Ba, Jimmy},
  journal={arXiv preprint arXiv:1412.6980},
  year={2014}
}

@inproceedings{10.1145/3615894.3628498,
  author = {Terashima, Haru and Tamura, Naoki and Shoji, Kazuyuki and Katayama, Shin and Urano, Kenta and Yonezawa, Takuro and Kawaguchi, Nobuo},
  title = {Human Mobility Prediction Challenge: Next Location Prediction using Spatiotemporal BERT},
  year = {2023},
  isbn = {9798400703560},
  publisher = {Association for Computing Machinery},
  address = {New York, NY, USA},
  url = {https://doi.org/10.1145/3615894.3628498},
  doi = {10.1145/3615894.3628498},
  booktitle = {Proceedings of the 1st International Workshop on the Human Mobility Prediction Challenge},
  pages = {1–6},
  numpages = {6},
  location = {Hamburg, Germany},
  series = {HuMob-Challenge '23}
}

@inproceedings{10.1145/3615894.3628499,
  author = {Solatorio, Aivin V.},
  title = {GeoFormer: Predicting Human Mobility using Generative Pre-trained Transformer (GPT)},
  year = {2023},
  isbn = {9798400703560},
  publisher = {Association for Computing Machinery},
  address = {New York, NY, USA},
  url = {https://doi.org/10.1145/3615894.3628499},
  doi = {10.1145/3615894.3628499},
  booktitle = {Proceedings of the 1st International Workshop on the Human Mobility Prediction Challenge},
  pages = {11–15},
  numpages = {5},
  location = {Hamburg, Germany},
  series = {HuMob-Challenge '23}
}

@inproceedings{10.1145/3615894.3628500,
  author = {Koyama, Ryo and Suzuki, Meisaku and Nakamura, Yusuke and Mimura, Tomohiro and Ishiguro, Shin},
  title = {Estimating future human trajectories from sparse time series data},
  year = {2023},
  isbn = {9798400703560},
  publisher = {Association for Computing Machinery},
  address = {New York, NY, USA},
  url = {https://doi.org/10.1145/3615894.3628500},
  doi = {10.1145/3615894.3628500},
  booktitle = {Proceedings of the 1st International Workshop on the Human Mobility Prediction Challenge},
  pages = {26–31},
  numpages = {6},
  location = {Hamburg, Germany},
  series = {HuMob-Challenge '23}
}

@inproceedings{10.1145/3681771.3699908,
  author = {Tang, Peizhi and Yang, Chuang and Xing, Tong and Xu, Xiaohang and Jiang, Renhe and Sezaki, Kaoru},
  title = {Instruction-Tuning Llama-3-8B Excels in City-Scale Mobility Prediction},
  year = {2024},
  isbn = {9798400711503},
  publisher = {Association for Computing Machinery},
  address = {New York, NY, USA},
  url = {https://doi.org/10.1145/3681771.3699908},
  doi = {10.1145/3681771.3699908},
  booktitle = {Proceedings of the 2nd ACM SIGSPATIAL International Workshop on Human Mobility Prediction Challenge},
  pages = {1–4},
  numpages = {4},
  location = {Atlanta, GA, USA},
  series = {HuMob'24}
}

\end{document}